\documentclass[conference]{IEEEtran}
\IEEEoverridecommandlockouts

\usepackage[T1]{fontenc}
\usepackage{newtxtext,newtxmath} 

\usepackage{cite}
\usepackage{algorithmic}
\usepackage{algorithm}
\usepackage{array}
\usepackage[caption=false,font=normalsize,labelfont=sf,textfont=sf]{subfig}
\usepackage{textcomp}
\usepackage{stfloats}
\usepackage{url}
\usepackage{bm}
\usepackage{color}
\usepackage[table,xcdraw]{xcolor}
\usepackage{multirow}
\usepackage{booktabs}
\usepackage{verbatim}
\usepackage{graphicx}
\usepackage{colortbl}
\usepackage{makecell}
\usepackage{tabularx}
\usepackage[most]{tcolorbox}
\usepackage{geometry}

\definecolor{lightred}{rgb}{1.0, 0.9, 0.9}
\definecolor{lightgreen}{rgb}{0.9, 1.0, 0.9}

\let\oldref\ref
\renewcommand{\ref}[1]{\textcolor{red}{\oldref{#1}}}

\usepackage[bookmarks=false]{hyperref}

\def\BibTeX{{\rm B\kern-.05em{\sc i\kern-.025em b}\kern-.08em
    T\kern-.1667em\lower.7ex\hbox{E}\kern-.125emX}}

\begin{document}

\title{Decoupling Semantics from Distortions: Multi-Scale Two-Stream Vision-Language Alignment for AI-Generated Image Quality Assessment}

\author{\IEEEauthorblockN{Zijie Meng}
\IEEEauthorblockA{Peking University, ymlf@stu.pku.edu.cn}
}

\maketitle

\begin{abstract}
Existing vision-language model (VLM)-based AI-generated image quality assessment (AIGIQA) methods suffer from a fundamental \emph{semantic-distortion dimensional conflict}: monolithic representations optimized for semantic discrimination inherently entangle compositional understanding with low-level perceptual sensitivity, rendering them blind to fine-grained quality degradations. We introduce MST-CLIPIQA, a multi-scale two-stream framework that achieves hierarchical vision-language alignment through explicit representational decoupling. Our architecture leverages dual CLIP encoders with complementary patch granularities—coarse-grained streams capture global semantic coherence while fine-grained streams preserve textural signatures and artifact patterns. An information bottleneck-inspired gated fusion mechanism performs adaptive cross-scale distillation, with optional cross-attention enabling prompt-anchored correspondence evaluation when generation prompts are available. Extensive experiments across five benchmarks establish new state-of-the-art results, achieving average improvements of 1.11\% SRCC on quality and 2.35\% SRCC on text-image correspondence prediction, while maintaining efficiency with only $\sim$0.8M trainable parameters. Our project is available at \url{https://github.com/YMlinfeng/MST-CLIPIQA}.
\end{abstract}


\section{Introduction}
\label{sec:introduction}

The rapid proliferation of AI-generated images (AIGIs) has introduced a fundamentally distinct quality assessment paradigm where perceptual fidelity, semantic authenticity, and text-image correspondence constitute inseparable quality dimensions that transcend conventional distortion-centric evaluation~\cite{li2023agiqa3k,wang2024aigciqa2023,zhang2024agiqa20k}. Vision-Language Models (VLMs) pre-trained on massive image-text corpora, most notably CLIP~\cite{radford2021learning}, have demonstrated remarkable zero-shot transfer capabilities by leveraging rich semantic priors~\cite{wang2023clipiqa,agnolucci2024qualityaware}. However, a fundamental \textbf{semantic-distortion dimensional conflict} persists: the representational geometry of these models is inherently optimized for high-level semantic discrimination rather than low-level perceptual sensitivity~\cite{kwon2024attiqa,tang2024clipagiqa}. This renders them systematically blind to fine-grained texture degradations, localized artifacts, and subtle generative anomalies that critically influence human quality judgments.

This dimensional conflict originates from a \emph{scale-semantic entanglement} inherent in single-scale visual representations. Human quality perception operates hierarchically: coarse-grained gestalt processing governs global coherence evaluation while fine-grained scrutiny detects local artifacts~\cite{chen2024topiq}. Yet monolithic feature extraction conflates these perceptual hierarchies into an undifferentiated embedding that sacrifices sensitivity at both ends of the spatial spectrum. The natural remedy of multi-scale feature extraction introduces a secondary challenge of \emph{cross-scale information redundancy}, where substantial representational overlap between adjacent granularities dilutes quality-discriminative signals when processed through conventional fusion strategies such as concatenation or learned linear combination~\cite{ke2021musiq}. This demands principled mechanisms for selective information distillation that contemporary architectures conspicuously lack. Furthermore, AIGIs uniquely exhibit \emph{generative semantic misalignment}, including anatomically implausible compositions, physically impossible configurations, and prompt-content discordance. Standard VLM embeddings trained on naturalistic image-text pairs fundamentally struggle to encode such degradations, necessitating explicit cross-modal reasoning that leverages generation prompts as privileged semantic anchors.

\begin{table}[t]
\caption{Performance comparison on authenticity score prediction.}
\centering
\resizebox{\columnwidth}{!}{
\begin{tabular}{l|cc|cc}
\toprule
\multirow{2}{*}{Method} & \multicolumn{2}{c|}{AIGCIQA2023} & \multicolumn{2}{c}{PKU-AIGIQA-4K} \\
\cmidrule(r){2-5}
 & SRCC & PLCC & SRCC & PLCC \\
\midrule
LinearityIQA \cite{LinearityIQA} & 0.6710 & 0.6640 & 0.6427 & 0.6305 \\
MUSIQ \cite{musiq} & 0.7185 & 0.7152 & 0.6348 & 0.5999 \\
HyperIQA \cite{hyperiqa} & 0.7060 & 0.6971 & 0.7093 & 0.6964 \\
StairIQA \cite{StairIQA} & 0.7352 & 0.7347 & 0.6835 & 0.6891 \\
MANIQA \cite{MANIQA} & 0.7829 & 0.7704 & 0.2559 & 0.2553 \\
LIQE \cite{LIQE} & 0.8010 & 0.7893 & 0.7823 & 0.7805 \\
AMFF-Net$^\sharp$ \cite{AMFF} & 0.7749 & 0.7643 & - & - \\
CLIP-AGIQA$^\sharp$ \cite{tang2025clip} & 0.7940 & 0.7797 & - & - \\
\midrule
MST-CLIPIQA & 0.8149 & 0.8026 & 0.7919 & 0.7905 \\
MST-CLIPIQA* & \textbf{0.8170} & \textbf{0.8049} & \textbf{0.7993} & \textbf{0.8003} \\
\bottomrule
\end{tabular}
}
\label{tab:auth}
\end{table}

To systematically address these intertwined challenges, we propose \textbf{MST-CLIPIQA}, a \underline{M}ulti-\underline{S}cale \underline{T}wo-stream framework that achieves hierarchical vision-language geometric alignment through explicit decoupling and subsequent reconciliation of semantic understanding with perceptual sensitivity. Our \emph{Multi-Scale Two-Stream Feature Extraction} (MSTFE) architecture constructs complementary processing pathways at distinct spatial granularities: coarse-grained streams capture global semantic coherence through efficient long-range contextual aggregation, while fine-grained streams preserve high-frequency textural signatures and localized artifact patterns. This design exploits the native patch-based tokenization flexibility of modern vision encoders without incurring explicit image pyramid construction overhead. Notably, our architecture remains agnostic to specific encoder implementations, enabling seamless integration with any lightweight backbone. The central innovation lies in our \emph{Gated Feature Fusion} (GFF) module, which implements \textbf{information bottleneck-guided cross-scale selective gating}. A learnable gating network dynamically computes per-dimensional selection coefficients that adaptively interpolate between multi-scale contributions, effectively compressing the joint representation to maximally preserve quality-predictive mutual information while filtering scale-specific redundancies. This achieves a principled balance between representational completeness and discriminative compactness with linear computational complexity. When generation prompts are available, a lightweight cross-attention mechanism further grounds quality predictions in text-image semantic correspondence.

\begin{figure*}[t]
\centering
\includegraphics[width=0.98\textwidth]{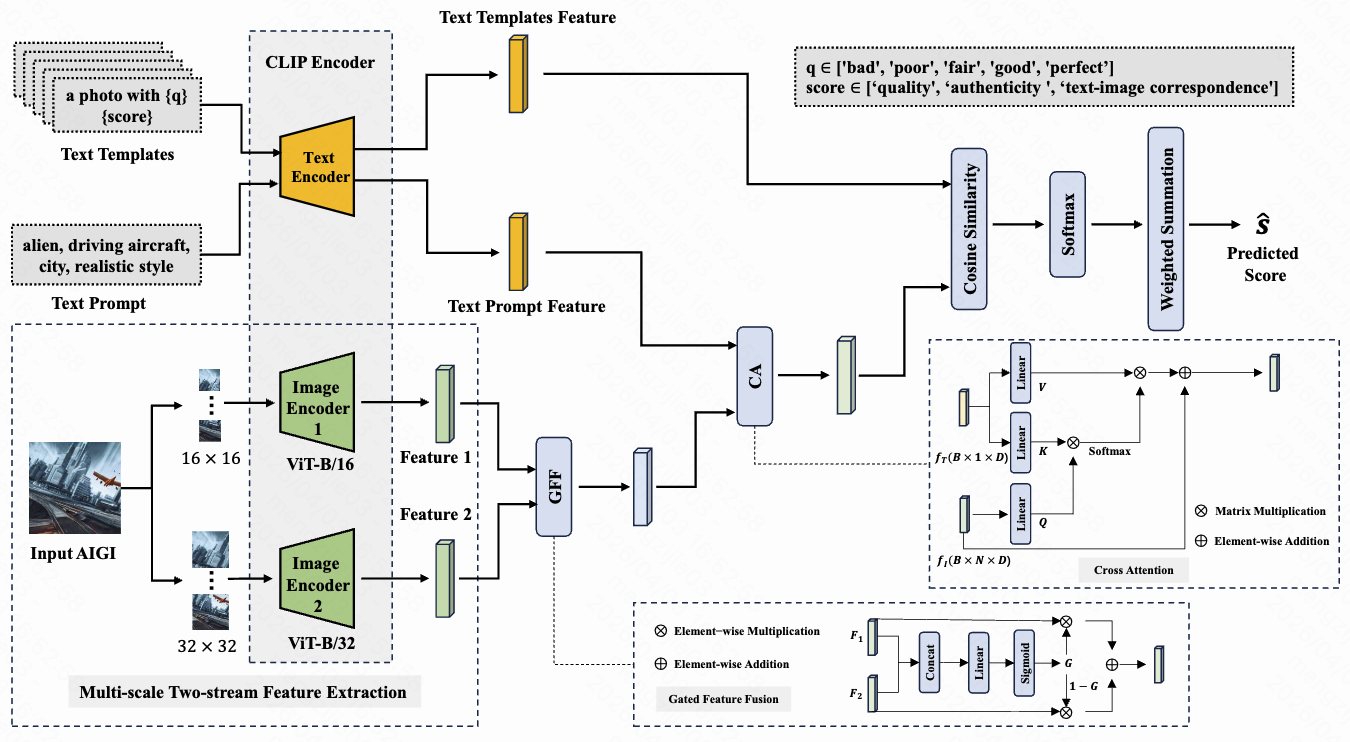}
\caption{The pipeline of the proposed CLIP-based multi-scale two-stream framework, which consists of five key modules: text feature extraction, multi-scale two-stream image feature extraction, gated feature fusion, text-image cross-attention, and score regression.}
\label{MSTCLIPIQA}
\end{figure*}

Extensive experiments across multiple AIGI quality benchmarks demonstrate that MST-CLIPIQA consistently achieves state-of-the-art performance, with particularly pronounced improvements on challenging cases involving subtle generative artifacts and complex prompt-image semantic relationships. Our contributions are fourfold:
\begin{itemize}
    \item We formalize the \textbf{semantic-distortion dimensional conflict} in VLM-based IQA and propose a multi-scale two-stream architecture (MSTFE) that explicitly decouples global semantic understanding from local perceptual sensitivity through complementary spatial granularity processing.
    
    \item We introduce \textbf{Gated Feature Fusion (GFF)}, an information bottleneck-inspired selective cross-scale gating mechanism that achieves adaptive quality-aware feature distillation with minimal computational overhead, substantially outperforming conventional fusion strategies.
    
    \item We present a \textbf{prompt-anchored cross-modal alignment} framework that elevates generation prompts to semantic quality references, enabling explicit text-image correspondence verification for AIGI-specific quality dimensions.
    
    \item We establish \textbf{new state-of-the-art results} on AGIQA-3K~\cite{li2023agiqa3k}, AIGCIQA2023~\cite{wang2024aigciqa2023}, and AIGIQA-20K~\cite{zhang2024agiqa20k}, demonstrating robust generalization across diverse generative paradigms from diffusion-based synthesis to GAN-generated imagery.
\end{itemize}



\begin{figure*}[!t]
\centering
\subfloat[LinearityIQA]{\includegraphics[width=0.24\textwidth]{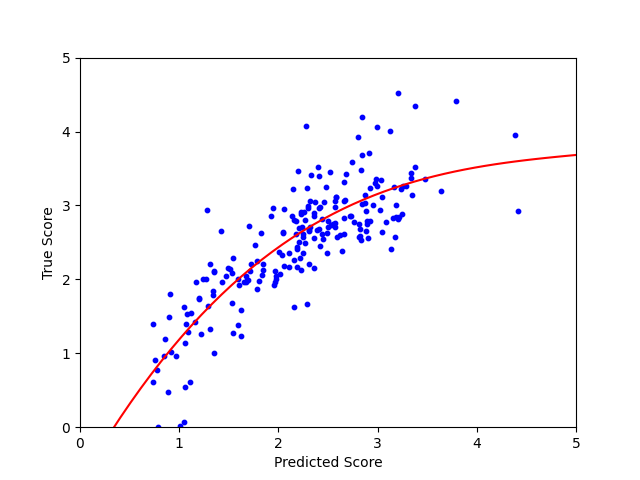}%
\label{LinearityIQA}}
\hfil
\subfloat[MUSIQ]{\includegraphics[width=0.24\textwidth]{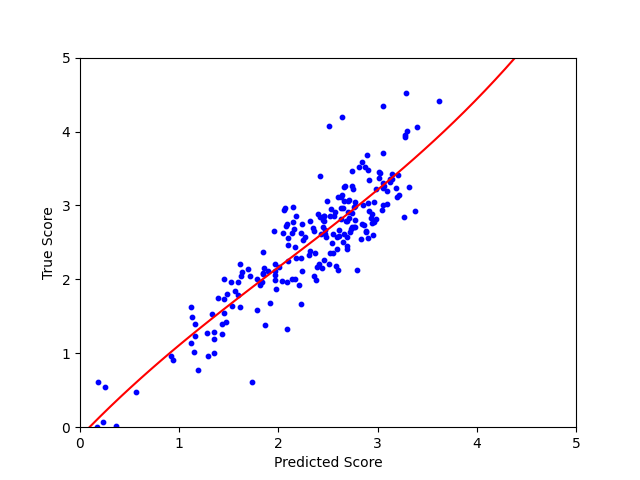}%
\label{MUSIQ}}
\hfil
\subfloat[HyperIQA]{\includegraphics[width=0.24\textwidth]{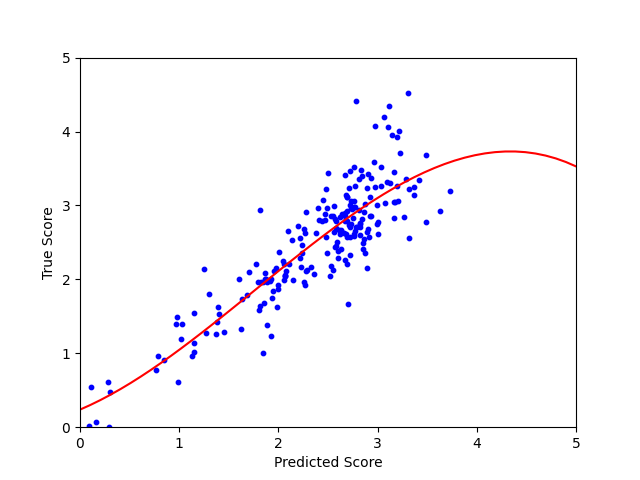}%
\label{HyperIQA}}
\hfil
\subfloat[StairIQA]{\includegraphics[width=0.24\textwidth]{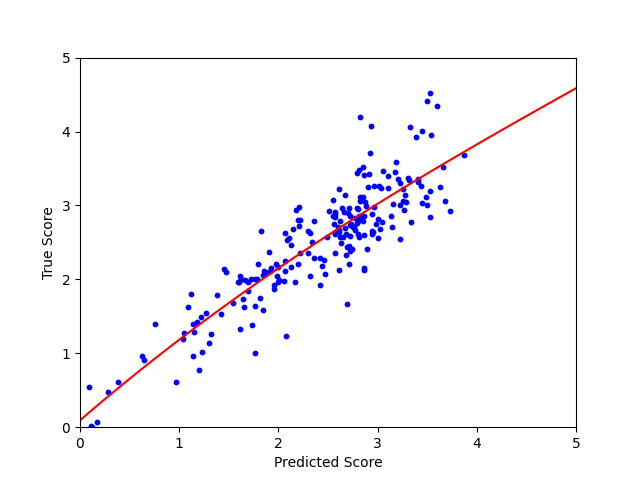}%
\label{StairIQA}}
\hfil
\subfloat[MANIQA]{\includegraphics[width=0.24\textwidth]{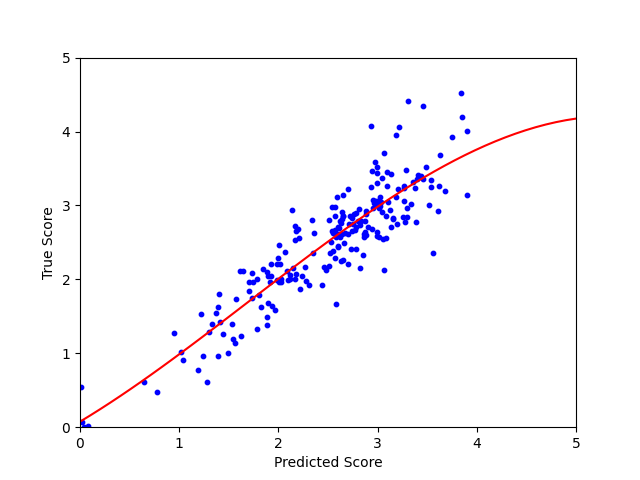}%
\label{MANIQA}}
\hfil
\subfloat[LIQE]{\includegraphics[width=0.24\textwidth]{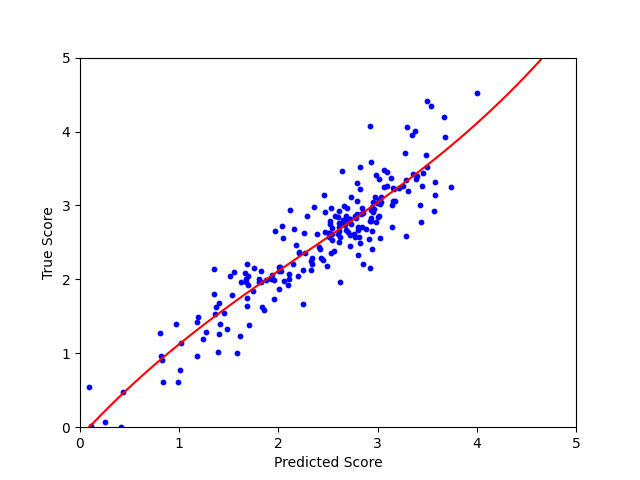}%
\label{LIQE}}
\hfil
\subfloat[MST-CLIPIQA]{\includegraphics[width=0.23\textwidth]{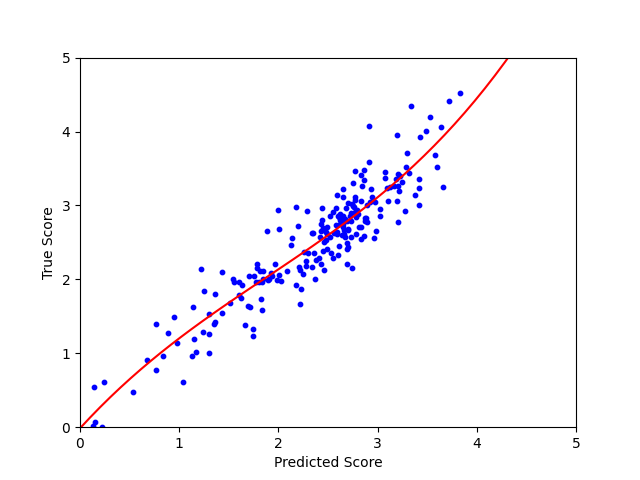}%
\label{MST-CLIPIQA}}
\hfil
\subfloat[MST-CLIPIQA*]{\includegraphics[width=0.24\textwidth]{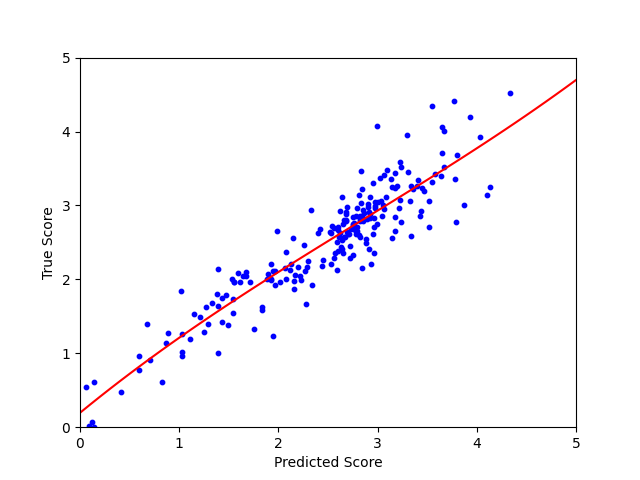}%
\label{MST-CLIPIQA*}}
\caption{Scatter plots of different IQA methods tested on the AGIQA-1K database. The curves are obtained by a three-order polynomial nonlinear fitting.}
\label{scatter}
\end{figure*}

\section{Method}
\label{sec:method}


\begin{table*}[t]
\caption{Comparison of multi-scale feature fusion methods.}
\centering
\small
\resizebox{\textwidth}{!}{
\begin{tabular}{l|cc|cc|cc|cc|cc}
\toprule
\multirow{2}{*}{Method} & \multicolumn{2}{c|}{AGIQA-1K} & \multicolumn{2}{c|}{AGIQA-3K} & \multicolumn{4}{c|}{AIGCIQA2023} & \multicolumn{2}{c}{\multirow{2}{*}{Average}}\\
\cmidrule(r){2-9}
 & \multicolumn{2}{c|}{Quality} & \multicolumn{2}{c|}{Quality} & \multicolumn{2}{c|}{Quality} & \multicolumn{2}{c|}{Authenticity} & \\
\cmidrule(r){2-11}
 & SRCC & PLCC & SRCC & PLCC & SRCC & PLCC & SRCC & PLCC & SRCC & PLCC\\
\midrule
Linear Interpolation & \textbf{0.9031} & 0.9094 & 0.9062 & 0.9244 & 0.8641 & 0.8731 & 0.8032 & 0.7937 & 0.8692 & 0.8752 \\
AdaptiveAvgPool & 0.8984 & 0.9090 & 0.9072 & 0.9265 & 0.8606 & 0.8780 & 0.8103 & 0.7996 & 0.8691 & 0.8783 \\
AdaptiveMaxPool & 0.8938 & 0.9112 & 0.9065 & 0.9256 & 0.8615 & 0.8767 & 0.8011 & 0.7889 & 0.8657 & 0.8756 \\
Cross Attention & 0.8944 & 0.9024 & 0.9053 & 0.9249 & 0.8561 & 0.8759 & 0.8033 & 0.7925 & 0.8648 & 0.8739 \\
\midrule
GFF (Ours) & 0.8990 & \textbf{0.9149} & \textbf{0.9091} & \textbf{0.9282} & \textbf{0.8701} & \textbf{0.8835} & \textbf{0.8149} & \textbf{0.8026} & \textbf{0.8733} & \textbf{0.8823} \\
\bottomrule
\end{tabular}
}
\label{tab:fusion}
\end{table*}

\begin{table*}[t]
\caption{Comparison of MSTFE architecture variants.}
\centering
\resizebox{\textwidth}{!}{
\begin{tabular}{l|cc|cc|cc|cc|cc}
\toprule
\multirow{2}{*}{Architecture} & \multicolumn{2}{c|}{AGIQA-1K} & \multicolumn{2}{c|}{AGIQA-3K} & \multicolumn{4}{c|}{AIGCIQA2023} & \multicolumn{2}{c}{\multirow{2}{*}{Average}}\\
\cmidrule(r){2-9}
 & \multicolumn{2}{c|}{Quality} & \multicolumn{2}{c|}{Quality} & \multicolumn{2}{c|}{Quality} & \multicolumn{2}{c|}{Authenticity} & \\
\cmidrule(r){2-11}
 & SRCC & PLCC & SRCC & PLCC & SRCC & PLCC & SRCC & PLCC & SRCC & PLCC\\
\midrule
MSTFE-1 & 0.8977 & 0.9071 & \textbf{0.9106} & \textbf{0.9289} & 0.8637 & 0.8780 & 0.8107 & 0.7960 & 0.8707 & 0.8775 \\
MSTFE-2 (Ours) & \textbf{0.8990} & \textbf{0.9149} & 0.9091 & 0.9282 & \textbf{0.8701} & \textbf{0.8835} & \textbf{0.8149} & \textbf{0.8026} & \textbf{0.8733} & \textbf{0.8823} \\
\bottomrule
\end{tabular}
}
\label{tab:arch}
\end{table*}

\subsection{Overview}
\label{sec:overview}

Given an input image $\mathbf{I} \in \mathbb{R}^{H \times W \times 3}$ with spatial resolution $H \times W$ and an optional generation prompt $\mathbf{T} = \{t_i\}_{i=1}^{L}$ of length $L$, MST-CLIPIQA predicts a perceptual quality score $\hat{q} \in \mathbb{R}$ through three cascaded stages. As illustrated in Fig.~\ref{MSTCLIPIQA}, our framework first employs Multi-Scale Two-Stream Feature Extraction (MSTFE) to construct scale-decoupled representations that resolve the semantic-distortion dimensional conflict (see Appendix A for theoretical analysis). Let $\mathcal{E}_\theta$ denote a frozen vision encoder parameterized by $\theta$ and $\mathcal{T}^{(s)}$ a granularity-specific tokenization operator with patch size $P_s$, the dual-stream features are obtained as:
\begin{align}
\mathbf{F}^{(c)} &= \mathcal{E}_\theta\bigl(\mathcal{T}^{(c)}(\mathbf{I})\bigr) \in \mathbb{R}^{N_c \times D} \label{eq:coarse_stream} \\
\mathbf{F}^{(f)} &= \mathcal{E}_\theta\bigl(\mathcal{T}^{(f)}(\mathbf{I})\bigr) \in \mathbb{R}^{N_f \times D} \label{eq:fine_stream}
\end{align}
where $N_s = HW / P_s^2$ denotes the token count for stream $s \in \{c, f\}$ and $D$ is the embedding dimension. The coarse-grained stream ($P_c > P_f$) captures global semantic coherence while the fine-grained stream preserves local textural details. Subsequently, Gated Feature Fusion (GFF) adaptively combines multi-scale representations through learned per-dimensional interpolation $\mathbf{z} = \text{GFF}(\mathbf{f}^{(c)}, \mathbf{f}^{(f)})$, filtering task-irrelevant redundancies. Finally, a lightweight regression head maps the fused representation to $\hat{q}$.

We denote our model as \textbf{MST-CLIPIQA} when operating without text prompts, where the fused visual features are compared against learnable quality-aware text templates. When generation prompts are available, we denote the variant as \textbf{MST-CLIPIQA*}, which additionally incorporates cross-modal attention to explicitly verify prompt-content alignment.

\begin{table*}[t]
\caption{Performance comparison on quality score prediction. Best results are in \textbf{bold}.}
\centering
\resizebox{\textwidth}{!}{
\begin{tabular}{l|cc|cc|cc|cc|cc}
\toprule
\multirow{2}{*}{Method} & \multicolumn{2}{c|}{AGIQA-1K} & \multicolumn{2}{c|}{AGIQA-3K} & \multicolumn{2}{c|}{AIGCIQA2023} & \multicolumn{2}{c|}{AIGIQA-20K} & \multicolumn{2}{c}{PKU-AIGIQA-4K}\\
\cmidrule(r){2-11}
 & SRCC & PLCC & SRCC & PLCC & SRCC & PLCC & SRCC & PLCC & SRCC & PLCC\\
\midrule
LinearityIQA \cite{LinearityIQA} & 0.8200 & 0.8578 & 0.8189 & 0.8309 & 0.7947 & 0.8057 & 0.7419 & 0.6838 & 0.6493 & 0.6056 \\
MUSIQ \cite{musiq} & 0.8506 & 0.8850 & 0.8338 & 0.8698 & 0.8261 & 0.8382 & 0.8344 & 0.8678 & 0.6801 & 0.6697 \\
HyperIQA \cite{hyperiqa} & 0.8534 & 0.8912 & 0.8495 & 0.8923 & 0.8159 & 0.8212 & 0.8174 & 0.8417 & 0.7144 & 0.7180 \\
StairIQA \cite{StairIQA} & 0.8640 & 0.8899 & 0.8543 & 0.8943 & 0.8313 & 0.8376 & 0.7911 & 0.8435 & 0.7247 & 0.7145 \\
MANIQA \cite{MANIQA} & 0.8804 & 0.9084 & 0.8916 & 0.9194 & 0.8412 & 0.8540 & 0.8553 & 0.8891 & 0.7800 & 0.7800 \\
LIQE \cite{LIQE} & 0.8927 & 0.9117 & 0.9009 & 0.9220 & 0.8608 & 0.8774 & 0.8655 & 0.8934 & 0.8030 & 0.8001 \\
MA-AGIQA$^\sharp$ \cite{MA-AGIQA} & - & - & 0.8939 & 0.9273 & - & - & 0.8644 & 0.9050 & - & - \\
\midrule
MST-CLIPIQA & 0.8990 & 0.9149 & \textbf{0.9091} & 0.9282 & \textbf{0.8701} & \textbf{0.8835} & 0.8803 & 0.9078 & \textbf{0.8289} & 0.8176 \\
MST-CLIPIQA* & \textbf{0.9091} & \textbf{0.9180} & 0.9085 & \textbf{0.9283} & 0.8608 & 0.8754 & \textbf{0.8936} & \textbf{0.9149} & 0.8261 & \textbf{0.8180} \\
\bottomrule
\end{tabular}
}
\label{tab:main_quality}
\end{table*}

\subsection{Multi-Scale Two-Stream Feature Extraction}
\label{sec:mstfe}

The MSTFE module exploits the native patch-based tokenization flexibility of vision-language encoders to construct scale-decoupled representations without explicit image pyramid construction. For an input image $\mathbf{I}$, the tokenization operator $\mathcal{T}^{(s)}$ with patch size $P_s$ partitions it into $N_s$ non-overlapping patches, each flattened and projected via $\mathbf{E}_p \in \mathbb{R}^{D \times (P_s^2 \cdot 3)}$ into $D$-dimensional embeddings. Together with a prepended \texttt{[CLS]} token $\mathbf{x}^{(s)}_{\texttt{cls}} \in \mathbb{R}^D$ and learnable positional embeddings $\mathbf{E}_{\text{pos}}^{(s)} \in \mathbb{R}^{(N_s+1) \times D}$, the input sequence $\mathbf{X}^{(s)} = [\mathbf{x}^{(s)}_{\texttt{cls}}; \mathbf{E}_p \cdot \text{Flatten}(\mathcal{T}^{(s)}(\mathbf{I})) + \mathbf{E}_{\text{pos}}^{(s)}] \in \mathbb{R}^{(N_s+1) \times D}$ is processed through $M$ stacked transformer blocks. The stream-specific global features are extracted from the final \texttt{[CLS]} representations:
\begin{equation}
\mathbf{f}^{(c)} = \mathcal{E}_\theta(\mathbf{X}^{(c)})[0], \quad \mathbf{f}^{(f)} = \mathcal{E}_\theta(\mathbf{X}^{(f)})[0] \in \mathbb{R}^D
\label{eq:cls_extraction}
\end{equation}
where $[\cdot][0]$ denotes extraction of the first token (\texttt{[CLS]}) from the output sequence.

The dual-stream design explicitly instantiates the hierarchical nature of human quality perception. Varying the patch size $P_s$ induces fundamentally different receptive field characteristics: larger patches ($P_c$) enforce spatial pooling that emphasizes compositional structure and semantic plausibility, directly addressing global coherence evaluation. Conversely, smaller patches ($P_f$) preserve fine-grained spatial locality, encoding textural patterns, edge sharpness, and localized artifact signatures that conventional VLM features systematically neglect. This transforms the entangled monolithic representation into a disentangled dual-component structure: $\mathbf{f}^{(c)}$ captures ``what is depicted'' while $\mathbf{f}^{(f)}$ encodes ``how well it is rendered.''

\subsection{Gated Feature Fusion}
\label{sec:gff}

Given the dual-stream features $\mathbf{f}^{(c)}, \mathbf{f}^{(f)} \in \mathbb{R}^D$, the GFF module learns to selectively combine multi-scale information through adaptive per-dimensional gating. We first concatenate the features and compute gate coefficients via a two-layer gating network with hidden dimension $D_h$:
\begin{equation}
\mathbf{g} = \sigma\Bigl(\mathbf{W}_g \cdot \text{ReLU}\bigl(\mathbf{W}_h [\mathbf{f}^{(c)}; \mathbf{f}^{(f)}] + \mathbf{b}_h\bigr) + \mathbf{b}_g\Bigr) \in [0,1]^D
\label{eq:gate_computation}
\end{equation}
where $\mathbf{W}_h \in \mathbb{R}^{D_h \times 2D}$ and $\mathbf{W}_g \in \mathbb{R}^{D \times D_h}$ are learnable projection matrices, $\mathbf{b}_h \in \mathbb{R}^{D_h}$ and $\mathbf{b}_g \in \mathbb{R}^D$ are bias terms, $[\cdot;\cdot]$ denotes concatenation, and $\sigma(\cdot)$ denotes the sigmoid function. The gate vector $\mathbf{g}$ determines the contribution of each scale at every feature dimension. The fused representation is then computed through element-wise interpolation:
\begin{equation}
\mathbf{z} = \mathbf{g} \odot \phi_c(\mathbf{f}^{(c)}) + (\mathbf{1} - \mathbf{g}) \odot \phi_f(\mathbf{f}^{(f)}) \in \mathbb{R}^D
\label{eq:gated_fusion}
\end{equation}
where $\odot$ denotes the Hadamard (element-wise) product, $\mathbf{1} \in \mathbb{R}^D$ is an all-ones vector, and $\phi_c(\cdot) = \mathbf{W}_c(\cdot) + \mathbf{b}_c$, $\phi_f(\cdot) = \mathbf{W}_f(\cdot) + \mathbf{b}_f$ with $\mathbf{W}_c, \mathbf{W}_f \in \mathbb{R}^{D \times D}$ are stream-specific affine projections that align feature distributions before fusion. When $g_i \rightarrow 1$, dimension $i$ predominantly reflects coarse-grained semantics; when $g_i \rightarrow 0$, fine-grained textural information dominates.

\textbf{MST-CLIPIQA (without prompts).} When generation prompts are unavailable, we compute quality scores by measuring the similarity between $\mathbf{z}$ and a set of learnable quality-level text embeddings $\{\mathbf{e}_k\}_{k=1}^{K}$ derived from templates (e.g., ``a photo of \{quality\} quality''), followed by softmax-weighted aggregation.

\textbf{MST-CLIPIQA* (with prompts).} When prompts $\mathbf{T}$ are available, we augment $\mathbf{z}$ with text-image correspondence through cross-modal attention: $\mathbf{z}' = \mathbf{z} + \alpha \cdot \text{CrossAttn}(\mathbf{z}, \mathcal{E}_{\text{text}}(\mathbf{T}))$, where $\mathcal{E}_{\text{text}}(\cdot)$ is the frozen text encoder and $\alpha$ is a learnable scalar initialized to zero for stable training. This enables explicit verification of prompt-content alignment. Detailed formulations of the cross-attention mechanism are provided in Appendix B.

The final quality score is obtained through a regression head with residual connection: $\hat{q} = \mathbf{w}_o^\top (\text{GELU}(\mathbf{W}_1 \mathbf{z}') + \mathbf{W}_2 \mathbf{z}') + b_o$, where $\mathbf{W}_1, \mathbf{W}_2 \in \mathbb{R}^{D \times D}$ are projection matrices, $\mathbf{w}_o \in \mathbb{R}^D$ and $b_o \in \mathbb{R}$ are output weights and bias. We optimize the model using a composite loss $\mathcal{L} = \mathcal{L}_{\text{MSE}} + \lambda \mathcal{L}_{\text{rank}}$ that jointly minimizes prediction error and enforces pairwise ranking consistency, where $\lambda$ controls the relative weight of the ranking term. The vision encoder $\mathcal{E}_\theta$ remains frozen throughout training, with only the lightweight fusion and regression parameters ($\sim$0.8M) being updated. Complete training details including hyperparameter settings are provided in Appendix C.

\section{Experiment}
\label{sec:experiment}

\subsection{Databases and Experiment Settings}
\label{sec:settings}

\noindent
\textbf{Databases.} We evaluate on five AIGIQA benchmarks: AGIQA-1K \cite{AGIQA1K} (1,080 images), AGIQA-3K \cite{AGIQA-3K} (2,982 images), AIGCIQA2023 \cite{aigciqa2023} (2,400 images), AIGIQA-20K \cite{AIGIQA20K} (14,000 training images used), and PKU-AIGIQA-4K \cite{PKUAIGIQA4K} (4,000 images). These databases provide MOS annotations for quality, authenticity, and text-image correspondence.

\noindent
\textbf{Evaluation Criteria.} We adopt Spearman rank correlation coefficient (SRCC) for monotonicity and Pearson linear correlation coefficient (PLCC) for accuracy.

\noindent
\textbf{Implementation Details.}
Experiments are conducted on NVIDIA A40 with PyTorch 1.11.0. We use Adam optimizer \cite{kingma2014adam} with learning rate $5 \times 10^{-6}$, weight decay $1 \times 10^{-3}$, and batch size 8. Datasets are split 4:1 for training and testing.

\subsection{Results and Analysis}
\label{sec:results}

\noindent
\textbf{Comparison with State-of-the-Art Methods.}
We compare against representative IQA methods including CNN-based approaches (LinearityIQA \cite{LinearityIQA}, MUSIQ \cite{musiq}, HyperIQA \cite{hyperiqa}), transformer-based methods (StairIQA \cite{StairIQA}, MANIQA \cite{MANIQA}), VLM-based approaches (LIQE \cite{LIQE}, CLIP-AGIQA \cite{tang2025clip}, MA-AGIQA \cite{MA-AGIQA}), and text-image matching methods (CLIPScore \cite{clipscore}, PickScore \cite{pickapic}, ImageReward \cite{imagereward}). Tables~\ref{tab:main_quality}, \ref{tab:auth}, and \ref{tab:corr} present performance comparisons on quality, authenticity, and text-image correspondence prediction, respectively. Methods marked with `$\sharp$' indicate results from original papers.

\begin{table}[t]
\caption{Performance comparison on text-image correspondence score prediction.}
\centering
\resizebox{\columnwidth}{!}{
\begin{tabular}{l|cc|cc|cc}
\toprule
\multirow{2}{*}{Method} & \multicolumn{2}{c|}{AGIQA-3K} & \multicolumn{2}{c|}{AIGCIQA2023} & \multicolumn{2}{c}{PKU-AIGIQA-4K}\\
\cmidrule(r){2-7}
 & SRCC & PLCC & SRCC & PLCC & SRCC & PLCC\\
\midrule
CLIPScore \cite{clipscore} & 0.5207 & 0.6409 & 0.2337 & 0.2483 & 0.1492 & 0.1969 \\
PickScore \cite{pickapic} & 0.6710 & 0.7252 & 0.5159 & 0.5136 & 0.5056 & 0.5682 \\
ImageReward \cite{imagereward} & 0.7297 & 0.7847 & 0.5870 & 0.5874 & 0.4832 & 0.5829 \\
LIQE \cite{LIQE} & 0.7638 & 0.8480 & 0.7529 & 0.7468 & 0.7796 & 0.7946 \\
AMFF-Net$^\sharp$ \cite{AMFF} & 0.7513 & 0.8476 & \textbf{0.7782} & 0.7638 & - & - \\
\midrule
MST-CLIPIQA & 0.7895 & 0.8666 & 0.7729 & 0.7615 & \textbf{0.8059} & 0.8141 \\
MST-CLIPIQA* & \textbf{0.8124} & \textbf{0.8817} & 0.7762 & \textbf{0.7696} & 0.8009 & \textbf{0.8174} \\
\bottomrule
\end{tabular}
}
\label{tab:corr}
\end{table}

As shown in Table~\ref{tab:main_quality}, both MST-CLIPIQA variants achieve state-of-the-art performance across all five benchmarks. Compared to the previous best method LIQE, MST-CLIPIQA achieves average improvements of 0.89\% SRCC and 0.93\% PLCC on quality prediction, while MST-CLIPIQA* further improves to 1.11\% SRCC and 0.98\% PLCC. For authenticity prediction (Table~\ref{tab:auth}), MST-CLIPIQA* outperforms LIQE by 1.12\% SRCC and 1.31\% PLCC on average, indicating that text-image alignment provides auxiliary cues for realism assessment. For correspondence prediction (Table~\ref{tab:corr}), MST-CLIPIQA* demonstrates particularly strong gains of 2.35\% SRCC and 1.99\% PLCC, significantly outperforming zero-shot text-image matching methods and validating the effectiveness of TICAM for prompt-aware evaluation. To provide intuitive comparison, Figure~\ref{scatter} presents scatter plots on AGIQA-1K, where MST-CLIPIQA* exhibits the tightest clustering around the diagonal, indicating superior prediction accuracy and consistency.

\noindent
\textbf{Effectiveness of Multi-Scale Feature Extraction.}
To validate the core contribution of MSTFE, we compare our dual-encoder design against single-encoder baselines. As shown in Table~\ref{tab:encoder}, employing complementary ViT-B/32 and ViT-B/16 encoders consistently outperforms either single encoder, with average improvements of 0.48\% SRCC and 0.72\% PLCC over the best single-scale baseline. This confirms that coarse-grained semantic features and fine-grained textural features provide complementary information for quality assessment.

\begin{table}[t]
\caption{Effects of image encoder configuration.}
\centering
\resizebox{\columnwidth}{!}{
\begin{tabular}{l|cc|cc|cc}
\toprule
\multirow{2}{*}{Encoder} & \multicolumn{2}{c|}{AGIQA-1K} & \multicolumn{2}{c|}{AIGCIQA2023} & \multicolumn{2}{c}{Average}\\
\cmidrule(r){2-7}
 & SRCC & PLCC & SRCC & PLCC & SRCC & PLCC\\
\midrule
ViT-B/16 & 0.8937 & 0.9021 & 0.8653 & 0.8789 & 0.8695 & 0.8760 \\
ViT-B/32 & 0.8980 & 0.9064 & 0.8613 & 0.8751 & 0.8685 & 0.8751 \\
\midrule
ViT-B/32 + ViT-B/16 & \textbf{0.8990} & \textbf{0.9149} & \textbf{0.8701} & \textbf{0.8835} & \textbf{0.8733} & \textbf{0.8823} \\
\bottomrule
\end{tabular}
}
\label{tab:encoder}
\end{table}




\noindent
\textbf{Effects of Multi-Scale Feature Fusion Methods.}
Table~\ref{tab:fusion} compares GFF against alternative fusion strategies including linear interpolation, adaptive pooling, and cross-attention. GFF achieves the best average performance (0.8733 SRCC, 0.8823 PLCC), as the learnable gating mechanism enables adaptive selection of task-relevant information from each scale.

\noindent
\textbf{Effects of MSTFE Architecture Variants.}
We compare two MSTFE architectures: MSTFE-1 with two complete CLIP models versus MSTFE-2 with dual image encoders sharing a single text encoder. Table~\ref{tab:arch} shows that MSTFE-2 achieves comparable or better performance with lower computational cost, validating that multi-scale visual features can be effectively aligned within a unified text embedding space.

\noindent
\textbf{Ablation Study.}
Table~\ref{tab:ablation} presents ablation results validating each component. Starting from the ViT-B/32 baseline, adding MSTFE improves SRCC by 0.10\%--1.02\% across different metrics. Incorporating cross-attention (CA) with text prompts yields larger gains of 0.30\%--2.64\%, particularly for correspondence prediction where prompt information is directly relevant. Combining both components achieves the best performance, demonstrating their complementary contributions.

\begin{table}[t]
\caption{Ablation study. MSTFE: multi-scale two-stream feature extraction; CA: cross-attention with text prompts.}
\centering
\resizebox{\columnwidth}{!}{
\begin{tabular}{cc|cc|cc|cc}
\toprule
\multirow{2}{*}{MSTFE} & \multirow{2}{*}{CA} & \multicolumn{2}{c|}{AGIQA-1K} & \multicolumn{2}{c|}{AGIQA-3K} & \multicolumn{2}{c}{AIGCIQA2023}\\
\cmidrule(r){3-8}
& & \multicolumn{2}{c|}{Quality} & \multicolumn{2}{c|}{Corresp.} & \multicolumn{2}{c}{Auth.} \\
\cmidrule(r){1-8}
 & & SRCC & PLCC & SRCC & PLCC & SRCC & PLCC \\
\midrule
- & - & 0.8980 & 0.9064 & 0.7793 & 0.8619 & 0.8070 & 0.7944 \\
$\checkmark$ & - & 0.8990 & 0.9149 & 0.7895 & 0.8666 & 0.8149 & 0.8026 \\
- & $\checkmark$ & 0.9010 & 0.9091 & 0.8057 & 0.8761 & 0.8109 & 0.7994 \\
$\checkmark$ & $\checkmark$ & \textbf{0.9091} & \textbf{0.9180} & \textbf{0.8124} & \textbf{0.8817} & \textbf{0.8170} & \textbf{0.8049} \\
\bottomrule
\end{tabular}
}
\label{tab:ablation}
\end{table}

\section{Conclusion}
\label{sec:conclusion}

We present MST-CLIPIQA, a multi-scale framework that addresses the semantic-distortion conflict in VLM-based AIGI quality assessment. By extracting global semantic features and fine-grained textural representations through dual-stream encoders with complementary patch sizes, our approach captures the hierarchical nature of human quality perception. The gated feature fusion enables adaptive integration of multi-scale information, while optional cross-attention with generation prompts enhances text-image correspondence evaluation. Experiments across five benchmarks demonstrate state-of-the-art performance on quality, authenticity, and correspondence prediction with only $\sim$0.8M trainable parameters.

\clearpage
\appendix

\section{Related Works}
\subsection{Image Quality Assessment (IQA)}
In recent years, the Contrastive Language-Image Pre-training (CLIP) model has attracted considerable attention in the field of Image Quality Assessment (IQA) due to its powerful vision-language alignment capabilities and rich prior knowledge learned from large-scale image-text pairs\cite{meng2025orpaint, meng2026make, liu2026omnidirector, meng2026argus, liu2025synpo, wei2025robust}. Wang \textit{et al.} \cite{CLIPIQA} pioneered the exploration of CLIP for assessing both the quality perception (i.e., the ``look'') and abstract perception (i.e., the ``feel'') of images, proposing CLIP-IQA which leverages effective prompt engineering and an antonym prompt pairing strategy (e.g., ``Good photo'' vs. ``Bad photo'') to harness CLIP's prior knowledge in a zero-shot manner, demonstrating that CLIP captures meaningful priors that generalize well to different perceptual assessments. Subsequently, Zhang \textit{et al.} \cite{LIQE} proposed LIQE (Learning Image Quality via Vision-Language Correspondence), a CLIP-based blind image quality assessment method that employs a multi-task learning paradigm to jointly learn three tasks---quality prediction, scene classification, and distortion type identification---by computing the joint probability from cosine similarities between visual and textual embeddings, thereby leveraging the correspondence between vision and language to predict image quality while benefiting from auxiliary task knowledge. 

With the rapid development of text-to-image generation models, recent research has increasingly focused on utilizing CLIP-based methods to evaluate AI-Generated Images (AIGIs), which present unique challenges including visual quality, authenticity, and text-image correspondence assessment. Zhou \textit{et al.} \cite{AMFF} introduced AMFF-Net (Adaptive Mixed-Scale Feature Fusion Network), a novel blind IQA framework that evaluates AGI quality from three dimensions---visual quality, authenticity, and text-image consistency---by employing a multi-scale input strategy inspired by the human visual system, utilizing an Adaptive Feature Fusion (AFF) block to adaptively fuse multi-scale features with learnable weights, and comparing semantic features from text and image encoders to assess text-to-image alignment. Qu \textit{et al.} \cite{IP-IQA} introduced IP-IQA, a CLIP-based dual-stream framework that simultaneously processes AI-generated images and their corresponding textual prompts, featuring an Image2Prompt incremental pretraining strategy to bridge AGI-style visual and textual modalities, and incorporating a cross-attention-based image-prompt fusion module along with a specially designed [QA] token to guide the model on quality-relevant aspects and enable effective image-text interaction. Unlike most existing methods that simply calculate similarity scores, they designed specialized modules to explicitly learn the deeper relationships between text prompts and input images. Peng \textit{et al.} \cite{IPCE_2024_CVPR} proposed IPCE, a CLIP-based AIGC image quality assessment method that emphasizes the correspondence between images and prompts by designing textual templates with five quality-related adverbs (e.g., ``badly'', ``poorly'', ``fairly'', ``well'', ``perfectly'') to represent different levels of image-prompt correlation, transforming the assessment into classification probabilities and subsequently into a precise regression task, achieving the first place in the image track of the NTIRE 2024 Quality Assessment for AI-Generated Content Challenge. Tang \textit{et al.} \cite{tang2025clip} proposed CLIP-AGIQA, a CLIP-based regression model for quality assessment of generated images that implements multi-category learnable prompts to fully utilize the textual knowledge encapsulated in CLIP, thereby enhancing prediction precision by moving beyond the limited contrastive similarity approach. 

However, despite these significant advances, existing methods have not fully explored the potential of CLIP in assessing the quality of AIGIs. First, most approaches utilize only a single pre-trained CLIP model for evaluation, which may limit the diversity of extracted features. Second, they primarily rely on calculating the similarity between text prompts and input images, which may constrain the model's ability to capture relationships beyond mere similarity measures. 


\section{Information Bottleneck Theory and Derivation}
\label{app:ib_derivation}

\subsection{Theoretical Background}

The Information Bottleneck framework addresses the fundamental problem of extracting relevant information from a source variable $\mathbf{X}$ about a target variable $Y$ while discarding irrelevant details. For our multi-scale feature fusion setting, we identify $\mathbf{X} \equiv \mathbf{F} = [\mathbf{f}^{(c)}; \mathbf{f}^{(f)}]$ as the concatenated multi-scale features and $Y \equiv q$ as the ground-truth quality score.

\paragraph{Mutual Information.} For continuous random variables, the mutual information $\mathcal{I}(\mathbf{X}; Y)$ quantifies the statistical dependence between $\mathbf{X}$ and $Y$:
\begin{equation}
\mathcal{I}(\mathbf{X}; Y) = \mathcal{H}(\mathbf{X}) - \mathcal{H}(\mathbf{X} | Y) = \mathcal{H}(Y) - \mathcal{H}(Y | \mathbf{X})
\label{eq:mutual_info_def}
\end{equation}
where $\mathcal{H}(\cdot)$ denotes differential entropy and $\mathcal{H}(\cdot | \cdot)$ denotes conditional entropy:
\begin{equation}
\mathcal{H}(\mathbf{X}) = -\mathbb{E}_{p(\mathbf{x})}\bigl[\log p(\mathbf{x})\bigr], \quad \mathcal{H}(\mathbf{X} | Y) = -\mathbb{E}_{p(\mathbf{x}, y)}\bigl[\log p(\mathbf{x} | y)\bigr]
\label{eq:entropy_def}
\end{equation}

\paragraph{IB Objective.} The IB principle seeks a compressed representation $\mathbf{Z}$ of $\mathbf{X}$ that preserves maximal information about $Y$ while minimizing information about $\mathbf{X}$ itself. This is formalized as a constrained optimization problem:
\begin{equation}
\min_{p(\mathbf{z}|\mathbf{x})} \; \mathcal{I}(\mathbf{X}; \mathbf{Z}) \quad \text{s.t.} \quad \mathcal{I}(\mathbf{Z}; Y) \geq \mathcal{I}_0
\label{eq:ib_constrained}
\end{equation}
Introducing a Lagrange multiplier $\beta^{-1}$ and reformulating as an unconstrained problem yields:
\begin{equation}
\mathcal{L}_{\text{IB}} = \mathcal{I}(\mathbf{X}; \mathbf{Z}) - \beta \cdot \mathcal{I}(\mathbf{Z}; Y) = -\beta \cdot \mathcal{I}(\mathbf{Z}; Y) + \mathcal{I}(\mathbf{Z}; \mathbf{X})
\label{eq:ib_lagrangian}
\end{equation}
which is equivalent to maximizing the objective in the main text.

\subsection{Variational Relaxation}

Direct optimization of Eq.~\eqref{eq:ib_lagrangian} is intractable due to the difficulty of estimating mutual information for high-dimensional continuous variables. Following the Variational Information Bottleneck (VIB) framework, we derive tractable upper and lower bounds.

\paragraph{Compression Term Upper Bound.} Using a variational marginal $r(\mathbf{z})$, we obtain:
\begin{align}
\mathcal{I}(\mathbf{X}; \mathbf{Z}) &= \mathbb{E}_{p(\mathbf{x})}\Bigl[\text{KL}\bigl(p(\mathbf{z}|\mathbf{x}) \| p(\mathbf{z})\bigr)\Bigr] \notag \\
&\leq \mathbb{E}_{p(\mathbf{x})}\Bigl[\text{KL}\bigl(p(\mathbf{z}|\mathbf{x}) \| r(\mathbf{z})\bigr)\Bigr]
\label{eq:compression_bound}
\end{align}
where $\text{KL}(\cdot \| \cdot)$ denotes the Kullback-Leibler divergence. The inequality follows from the non-negativity of KL divergence.

\paragraph{Relevance Term Lower Bound.} Using a variational decoder $q(y|\mathbf{z})$:
\begin{align}
\mathcal{I}(\mathbf{Z}; Y) &= \mathcal{H}(Y) - \mathcal{H}(Y | \mathbf{Z}) \notag \\
&\geq \mathcal{H}(Y) + \mathbb{E}_{p(\mathbf{z}, y)}\bigl[\log q(y | \mathbf{z})\bigr]
\label{eq:relevance_bound}
\end{align}
Combining these bounds yields the VIB objective:
\begin{equation}
\mathcal{L}_{\text{VIB}} = \mathbb{E}_{p(\mathbf{x})}\Bigl[\text{KL}\bigl(p(\mathbf{z}|\mathbf{x}) \| r(\mathbf{z})\bigr)\Bigr] - \beta \cdot \mathbb{E}_{p(\mathbf{x}, y)}\mathbb{E}_{p(\mathbf{z}|\mathbf{x})}\bigl[\log q(y | \mathbf{z})\bigr]
\label{eq:vib_objective}
\end{equation}

\subsection{Connection to Gated Feature Fusion}

Our GFF module implements a deterministic approximation to the VIB framework tailored for multi-scale feature fusion.

\paragraph{Deterministic Encoder.} Rather than learning a stochastic encoder $p(\mathbf{z}|\mathbf{x})$, we employ a deterministic gating function:
\begin{equation}
\mathbf{z} = f_\text{GFF}(\mathbf{F}; \Theta_g) = \mathbf{g}(\mathbf{F}) \odot \phi_c(\mathbf{f}^{(c)}) + (\mathbf{1} - \mathbf{g}(\mathbf{F})) \odot \phi_f(\mathbf{f}^{(f)})
\label{eq:deterministic_encoder}
\end{equation}
This can be viewed as a delta distribution encoder: $p(\mathbf{z}|\mathbf{x}) = \delta\bigl(\mathbf{z} - f_\text{GFF}(\mathbf{F})\bigr)$.

\paragraph{Implicit Compression.} The gating mechanism achieves implicit compression through selective dimension-wise interpolation. Let $g_i \in [0,1]$ denote the $i$-th gate value. The effective information content of dimension $i$ is bounded by:
\begin{equation}
\mathcal{I}(z_i; F_i) \leq g_i \cdot \mathcal{I}\bigl(\phi_c(f^{(c)}_i); F_i\bigr) + (1-g_i) \cdot \mathcal{I}\bigl(\phi_f(f^{(f)}_i); F_i\bigr)
\label{eq:implicit_compression}
\end{equation}
When the gate learns to select the more quality-relevant scale for each dimension, it effectively implements task-aware information routing that discards scale-specific redundancies.

\paragraph{Sparsity-Inducing Regularization.} To encourage explicit compression behavior, we can optionally augment the training objective with a gate sparsity penalty:
\begin{equation}
\mathcal{L}_{\text{sparse}} = \frac{1}{D} \sum_{i=1}^{D} \min\bigl(g_i, 1 - g_i\bigr)
\label{eq:sparsity_loss}
\end{equation}
This term encourages gate values toward the extremes (0 or 1), promoting hard selection that maximizes compression. In practice, we find this regularization unnecessary as the MSE and ranking losses provide sufficient gradient signal for meaningful gate specialization.

\section{Cross-Modal Attention Mechanism}
\label{app:cross_attention}

This appendix details the optional cross-modal attention module that enables prompt-anchored quality assessment when generation prompts are available.

\subsection{Motivation}

AI-generated images uniquely exhibit quality dimensions that require semantic reasoning beyond pure visual analysis. Specifically, \emph{prompt-content misalignment} occurs when the generated image fails to faithfully represent the textual generation prompt. Standard visual features cannot detect such misalignment without access to the prompt itself. Our cross-modal attention mechanism addresses this by computing explicit text-image correspondence signals.

\subsection{Architecture}

\paragraph{Text Encoding.} The generation prompt $\mathbf{T} = \{t_1, \ldots, t_L\}$ is processed through the frozen CLIP text encoder to obtain a sequence of contextualized token embeddings:
\begin{equation}
\mathbf{E}_t = \text{TextEncoder}(\mathbf{T}) = [\mathbf{e}_1, \ldots, \mathbf{e}_L] \in \mathbb{R}^{L \times D}
\label{eq:text_encoding}
\end{equation}
The \texttt{[EOS]} token embedding serves as the global text representation: $\mathbf{e}_t = \mathbf{e}_L \in \mathbb{R}^D$.

\paragraph{Cross-Attention Formulation.} We employ scaled dot-product attention with the fused visual representation $\mathbf{z}$ as the query and text embeddings as keys and values. First, we compute query, key, and value projections:
\begin{equation}
\mathbf{Q} = \mathbf{W}_Q \mathbf{z} \in \mathbb{R}^{D_k}, \quad \mathbf{K} = \mathbf{W}_K \mathbf{E}_t^\top \in \mathbb{R}^{D_k \times L}, \quad \mathbf{V} = \mathbf{W}_V \mathbf{E}_t^\top \in \mathbb{R}^{D_v \times L}
\label{eq:qkv_projection}
\end{equation}
where $\mathbf{W}_Q \in \mathbb{R}^{D_k \times D}$, $\mathbf{W}_K \in \mathbb{R}^{D_k \times D}$, $\mathbf{W}_V \in \mathbb{R}^{D_v \times D}$ are learnable projection matrices.

The attention weights and output are computed as:
\begin{equation}
\mathbf{A} = \text{softmax}\Bigl(\frac{\mathbf{Q}^\top \mathbf{K}}{\sqrt{D_k}}\Bigr) \in \mathbb{R}^L, \quad \mathbf{o} = \mathbf{V} \mathbf{A} \in \mathbb{R}^{D_v}
\label{eq:attention_computation}
\end{equation}

\paragraph{Multi-Head Extension.} For enhanced representational capacity, we employ multi-head attention with $H$ parallel attention heads:
\begin{equation}
\text{MultiHead}(\mathbf{z}, \mathbf{E}_t) = \mathbf{W}_O \cdot \text{Concat}\bigl(\mathbf{o}_1, \ldots, \mathbf{o}_H\bigr)
\label{eq:multihead}
\end{equation}
where each head $h$ operates with independent projections $\mathbf{W}_Q^{(h)}, \mathbf{W}_K^{(h)}, \mathbf{W}_V^{(h)}$ of reduced dimension $D_k/H$ and $D_v/H$, and $\mathbf{W}_O \in \mathbb{R}^{D \times D_v}$ is the output projection.

\paragraph{Residual Integration.} The cross-modal enhanced representation incorporates the attention output through a learnable residual connection:
\begin{equation}
\mathbf{z}' = \mathbf{z} + \alpha \cdot \text{LayerNorm}\bigl(\text{MultiHead}(\mathbf{z}, \mathbf{E}_t)\bigr)
\label{eq:residual_integration}
\end{equation}
where $\alpha$ is a learnable scalar initialized to zero. This zero initialization ensures that the model initially behaves identically to the prompt-free variant, with cross-modal information gradually incorporated as training progresses.

\subsection{Semantic Correspondence Interpretation}

The attention weights $\mathbf{A} \in \mathbb{R}^L$ provide interpretable signals about which prompt tokens the model considers most relevant for quality assessment. High attention on specific tokens (e.g., object names, attributes) indicates that the model is verifying whether these semantic elements are faithfully represented in the image. This enables post-hoc analysis of prompt-content alignment failures that degrade perceived quality.

\section{Implementation and Training Details}
\label{app:implementation}

This appendix provides comprehensive implementation details to ensure reproducibility.

\subsection{Architecture Specifications}

\paragraph{Vision Encoder.} We employ the CLIP ViT-B/32 vision encoder as our default backbone, which consists of $M = 12$ transformer blocks with embedding dimension $D = 512$. The encoder processes $224 \times 224$ images with the default patch size of $32 \times 32$. All encoder parameters remain frozen throughout training.

\paragraph{Multi-Scale Configuration.} For the dual-stream architecture, we configure:
\begin{equation}
\begin{cases}
\text{Coarse stream:} & P_c = 32, \; N_c = 49 \\
\text{Fine stream:} & P_f = 16, \; N_f = 196
\end{cases}
\label{eq:scale_config}
\end{equation}
The fine-grained stream requires interpolation of positional embeddings from the original $7 \times 7$ grid to a $14 \times 14$ grid, implemented via bicubic interpolation.

\paragraph{Gated Feature Fusion.} The gating network employs:
\begin{equation}
\begin{cases}
\text{Input dimension:} & 2D = 1024 \\
\text{Hidden dimension:} & D_h = 256 \\
\text{Output dimension:} & D = 512
\end{cases}
\label{eq:gff_config}
\end{equation}
Stream-specific projections $\phi_c, \phi_f$ are implemented as single linear layers without bias terms.

\paragraph{Regression Head.} The quality regression MLP uses:
\begin{equation}
\begin{cases}
\text{Hidden dimension:} & D_r = 128 \\
\text{Activation:} & \text{GELU} \\
\text{Dropout:} & 0.1 \text{ (training only)}
\end{cases}
\label{eq:regression_config}
\end{equation}

\paragraph{Cross-Modal Attention.} When enabled:
\begin{equation}
\begin{cases}
\text{Number of heads:} & H = 8 \\
\text{Key/Query dimension:} & D_k = 512 \\
\text{Value dimension:} & D_v = 512
\end{cases}
\label{eq:attention_config}
\end{equation}

\subsection{Training Protocol}

\paragraph{Optimization.} We employ the Adam optimizer with the following hyperparameters:
\begin{equation}
\begin{cases}
\text{Learning rate:} & \eta = 1 \times 10^{-4} \\
\text{Weight decay:} & \lambda_w = 1 \times 10^{-2} \\
\text{Betas:} & (\beta_1, \beta_2) = (0.9, 0.999) \\
\text{Epsilon:} & \epsilon_{\text{Adam}} = 1 \times 10^{-8}
\end{cases}
\label{eq:optimizer_config}
\end{equation}

\paragraph{Learning Rate Schedule.} We employ a cosine annealing schedule with linear warmup:
\begin{equation}
\eta_t = \begin{cases}
\eta \cdot \frac{t}{T_{\text{warmup}}} & \text{if } t \leq T_{\text{warmup}} \\
\eta_{\min} + \frac{1}{2}(\eta - \eta_{\min})\Bigl(1 + \cos\bigl(\frac{t - T_{\text{warmup}}}{T - T_{\text{warmup}}} \pi\bigr)\Bigr) & \text{otherwise}
\end{cases}
\label{eq:lr_schedule}
\end{equation}
where $T_{\text{warmup}} = 5$ epochs, $T = 50$ total epochs, and $\eta_{\min} = 1 \times 10^{-6}$.

\paragraph{Loss Hyperparameters.} The composite loss function uses:
\begin{equation}
\lambda = 0.1, \quad \epsilon = 0.05
\label{eq:loss_hyperparams}
\end{equation}
where $\lambda$ balances MSE and ranking losses, and $\epsilon$ is the ranking margin.

\paragraph{Data Augmentation.} Training images undergo:
\begin{equation}
\begin{cases}
\text{Random horizontal flip:} & p = 0.5 \\
\text{Random resized crop:} & \text{scale} \in [0.8, 1.0], \; \text{ratio} \in [0.9, 1.1] \\
\text{Color jitter:} & \text{brightness} = 0.1, \; \text{contrast} = 0.1
\end{cases}
\label{eq:augmentation}
\end{equation}
Validation and test images use center crop only.

\paragraph{Batch Configuration.} Training uses a batch size of $|\mathcal{B}| = 32$ per GPU. For multi-GPU training, we employ synchronized batch normalization equivalents where applicable.

\subsection{Inference Protocol}

During inference, the model operates in evaluation mode with all stochastic elements (dropout) disabled. Input images are resized to $224 \times 224$ using bicubic interpolation followed by center cropping. No test-time augmentation is employed. The predicted quality score $\hat{q}$ is output directly without post-processing.

\subsection{Computational Requirements}

\paragraph{Parameter Count.} The trainable parameters of MST-CLIPIQA total approximately 0.8M, distributed as:
\begin{equation}
\begin{cases}
\text{Gating network } (\psi_\phi): & \sim 0.4\text{M} \\
\text{Stream projections } (\phi_c, \phi_f): & \sim 0.3\text{M} \\
\text{Regression head:} & \sim 0.1\text{M}
\end{cases}
\label{eq:param_count}
\end{equation}
The frozen CLIP ViT-B/32 encoder contains 86M parameters.

\paragraph{Inference Speed.} On a single NVIDIA RTX 3090 GPU:
\begin{equation}
\begin{cases}
\text{Throughput:} & \sim 180 \text{ images/second} \\
\text{Latency:} & \sim 5.6 \text{ ms/image}
\end{cases}
\label{eq:inference_speed}
\end{equation}

\paragraph{Training Time.} Full training on AGIQA-3K requires approximately 2 hours on a single A40 GPU.

\bibliographystyle{IEEEbib}
\bibliography{icme2026references}

@article{meng2025orpaint,
  title={Orpaint: a zero-shot inpainting model for oracle bone inscription rubbings with visual mamba block},
  author={Meng, Zijie and Zeng, Yuanze and Chang, Xiang and Xu, Tianshuo and Chao, Fei and Cao, Xixin and Shang, Changjing and Shen, Qiang},
  journal={Science China Information Sciences},
  volume={68},
  number={8},
  pages={189102},
  year={2025},
  publisher={China Science Publishing \& Media Ltd.}
}

@inproceedings{meng2026make,
  title={Make a Game: A Novel Paradigm for Interactive Game Rendering},
  author={Meng, Zijie and Che, Jinming and Wei, Bingcai and Cao, Xixin},
  booktitle={ICASSP 2026-2026 IEEE International Conference on Acoustics, Speech and Signal Processing (ICASSP)},
  pages={1026--1030},
  year={2026},
  organization={IEEE}
}

@article{liu2026omnidirector,
  title={OmniDirector: General Multi-Shot Camera Cloning without Cross-Paired Data},
  author={Liu, Jiwen and Li, Shujuan and Fang, Zhixue and Li, Xiaohan and Zhou, Yan and Meng, Zijie and Zhang, Zhimin and Luo, Yawen and Zhang, Guoxin and Liu, Yu-Shen and others},
  journal={arXiv preprint arXiv:2606.13432},
  year={2026}
}

@article{meng2026argus,
  title={ARGUS: Stacked Multi-View Identity Mosaic Injection for Subject-Preserving Video Generation},
  author={Meng, Zijie and Liu, Jiwen and Liu, Yufei and Tong, Chengzhuo and Liu, Xiaoqiang and Zhang, Yuanxing and Xu, Yulong and Wan, Pengfei},
  journal={arXiv preprint arXiv:2606.11670},
  year={2026}
}

@inproceedings{liu2025synpo,
  title={SynPo: Boosting Training-Free Few-Shot Medical Segmentation via High-Quality Negative Prompts},
  author={Liu, Yufei and Xiao, Haoke and Chai, Jiaxing and Zhang, Yongcun and Wang, Rong and Meng, Zijie and Luo, Zhiming},
  booktitle={International Conference on Medical Image Computing and Computer-Assisted Intervention},
  pages={594--603},
  year={2025},
  organization={Springer}
}

@inproceedings{wei2025robust,
  title={Robust Single Image Sand Removal by Leveraging Uncertainty-aware SAM Priors and Prompt Learning with Refined Perceptual Loss},
  author={Wei, Bingcai and Liu, Hui and Qian, Chuang and Li, Zijian and Wu, Wangyu and Meng, Zijie},
  booktitle={Proceedings of the 33rd ACM International Conference on Multimedia},
  pages={4932--4941},
  year={2025}
}

@inproceedings{zhang2024agiqa20k,
  title={Aigiqa-20k: A large database for ai-generated image quality assessment},
  author={Li, Chunyi and Kou, Tengchuan and Gao, Yixuan and Cao, Yuqin and Sun, Wei and Zhang, Zicheng and Zhou, Yingjie and Zhang, Zhichao and Zhang, Weixia and Wu, Haoning and others},
  booktitle={Proceedings of the IEEE/CVF Conference on Computer Vision and Pattern Recognition},
  pages={6327--6336},
  year={2024}
}

@inproceedings{wang2024aigciqa2023,
  title={Aigciqa2023: A large-scale image quality assessment database for ai generated images: from the perspectives of quality, authenticity and correspondence},
  author={Wang, Jiarui and Duan, Huiyu and Liu, Jing and Chen, Shi and Min, Xiongkuo and Zhai, Guangtao},
  booktitle={CAAI International Conference on Artificial Intelligence},
  pages={46--57},
  year={2023},
  organization={Springer}
}

@article{li2023agiqa3k,
  title={Agiqa-3k: An open database for ai-generated image quality assessment},
  author={Li, Chunyi and Zhang, Zicheng and Wu, Haoning and Sun, Wei and Min, Xiongkuo and Liu, Xiaohong and Zhai, Guangtao and Lin, Weisi},
  journal={IEEE Transactions on Circuits and Systems for Video Technology},
  volume={34},
  number={8},
  pages={6833--6846},
  year={2023},
  publisher={IEEE}
}

@inproceedings{kwon2024attiqa,
  title={ATTIQA: Generalizable image quality feature extractor using attribute-aware pretraining},
  author={Kwon, Daekyu and Kim, Dongyoung and Ki, Sehwan and Jo, Younghyun and Lee, Hyong-Euk and Kim, Seon Joo},
  booktitle={Proceedings of the Asian Conference on Computer Vision},
  pages={4526--4543},
  year={2024}
}

@inproceedings{tang2024clipagiqa,
  title={CLIP-AGIQA: boosting the performance of ai-generated image quality assessment with clip},
  author={Tang, Zhenchen and Wang, Zichuan and Peng, Bo and Dong, Jing},
  booktitle={International Conference on Pattern Recognition},
  pages={48--61},
  year={2024},
  organization={Springer}
}

@inproceedings{ke2021musiq,
  title={Musiq: Multi-scale image quality transformer},
  author={Ke, Junjie and Wang, Qifei and Wang, Yilin and Milanfar, Peyman and Yang, Feng},
  booktitle={Proceedings of the IEEE/CVF international conference on computer vision},
  pages={5148--5157},
  year={2021}
}

@article{chen2024topiq,
  title={Topiq: A top-down approach from semantics to distortions for image quality assessment},
  author={Chen, Chaofeng and Mo, Jiadi and Hou, Jingwen and Wu, Haoning and Liao, Liang and Sun, Wenxiu and Yan, Qiong and Lin, Weisi},
  journal={IEEE Transactions on Image Processing},
  volume={33},
  pages={2404--2418},
  year={2024},
  publisher={IEEE}
}

@article{agnolucci2024qualityaware,
  title={Quality-aware image-text alignment for opinion-unaware image quality assessment},
  author={Agnolucci, Lorenzo and Galteri, Leonardo and Bertini, Marco},
  journal={arXiv preprint arXiv:2403.11176},
  year={2024}
}

@inproceedings{wang2023clipiqa,
  title={Exploring clip for assessing the look and feel of images},
  author={Wang, Jianyi and Chan, Kelvin CK and Loy, Chen Change},
  booktitle={Proceedings of the AAAI conference on artificial intelligence},
  volume={37},
  number={2},
  pages={2555--2563},
  year={2023}
}

@inproceedings{radford2021learning,
  title={Learning transferable visual models from natural language supervision},
  author={Radford, Alec and Kim, Jong Wook and Hallacy, Chris and Ramesh, Aditya and Goh, Gabriel and Agarwal, Sandhini and Sastry, Girish and Askell, Amanda and Mishkin, Pamela and Clark, Jack and others},
  booktitle={International conference on machine learning},
  pages={8748--8763},
  year={2021},
  organization={PmLR}
}

@inproceedings{musiq,
  title={Musiq: Multi-scale image quality transformer},
  author={Ke, Junjie and Wang, Qifei and Wang, Yilin and Milanfar, Peyman and Yang, Feng},
  booktitle={Proceedings of the IEEE/CVF international conference on computer vision},
  pages={5148--5157},
  year={2021}
}

@article{MANIQA,
  title={MANIQA: Multi-dimension Attention Network for No-Reference Image Quality Assessment},
  author={Sidi Yang and Tianhe Wu and Shu Shi and Shan Gong and Ming Cao and Jiahao Wang and Yujiu Yang},
  journal={2022 IEEE/CVF Conference on Computer Vision and Pattern Recognition Workshops (CVPRW)},
  year={2022},
  pages={1190-1199},
  url={https://api.semanticscholar.org/CorpusID:248240148}
}

@inproceedings{LIQE,
  title={Blind image quality assessment via vision-language correspondence: A multitask learning perspective},
  author={Zhang, Weixia and Zhai, Guangtao and Wei, Ying and Yang, Xiaokang and Ma, Kede},
  booktitle={Proceedings of the IEEE/CVF conference on computer vision and pattern recognition},
  pages={14071--14081},
  year={2023}
}

@inproceedings{LinearityIQA,
  title={Norm-in-norm loss with faster convergence and better performance for image quality assessment},
  author={Li, Dingquan and Jiang, Tingting and Jiang, Ming},
  booktitle={Proceedings of the 28th ACM International conference on multimedia},
  pages={789--797},
  year={2020}
}

@INPROCEEDINGS{StairIQA,
  author={Sun, Wei and Duan, Huiyu and Min, Xiongkuo and Chen, Li and Zhai, Guangtao},
  booktitle={2022 IEEE International Symposium on Broadband Multimedia Systems and Broadcasting (BMSB)}, 
  title={Blind Quality Assessment for in-the-Wild Images via Hierarchical Feature Fusion Strategy}, 
  year={2022},
  volume={},
  number={},
  pages={01-06},
  doi={10.1109/BMSB55706.2022.9828590}}

@INPROCEEDINGS{hyperiqa,
  author={Su, Shaolin and Yan, Qingsen and Zhu, Yu and Zhang, Cheng and Ge, Xin and Sun, Jinqiu and Zhang, Yanning},
  booktitle={2020 IEEE/CVF Conference on Computer Vision and Pattern Recognition (CVPR)}, 
  title={Blindly Assess Image Quality in the Wild Guided by a Self-Adaptive Hyper Network}, 
  year={2020},
  volume={},
  number={},
  pages={3664-3673},
  doi={10.1109/CVPR42600.2020.00372}}

@article{clipscore,
  title={Clipscore: A reference-free evaluation metric for image captioning},
  author={Hessel, Jack and Holtzman, Ari and Forbes, Maxwell and Bras, Ronan Le and Choi, Yejin},
  journal={arXiv preprint arXiv:2104.08718},
  year={2021}
}

@misc{pickapic,
      title={Pick-a-Pic: An Open Dataset of User Preferences for Text-to-Image Generation}, 
      author={Yuval Kirstain and Adam Polyak and Uriel Singer and Shahbuland Matiana and Joe Penna and Omer Levy},
      year={2023},
      eprint={2305.01569},
      archivePrefix={arXiv},
      primaryClass={cs.CV}
}

@misc{imagereward,
      title={ImageReward: Learning and Evaluating Human Preferences for Text-to-Image Generation}, 
      author={Jiazheng Xu and Xiao Liu and Yuchen Wu and Yuxuan Tong and Qinkai Li and Ming Ding and Jie Tang and Yuxiao Dong},
      year={2023},
      eprint={2304.05977},
      archivePrefix={arXiv},
      primaryClass={cs.CV}
}

@article{AMFF,
  title={Adaptive mixed-scale feature fusion network for blind AI-generated image quality assessment},
  author={Zhou, Tianwei and Tan, Songbai and Zhou, Wei and Luo, Yu and Wang, Yuan-Gen and Yue, Guanghui},
  journal={IEEE Transactions on Broadcasting},
  year={2024},
  publisher={IEEE}
}

@InProceedings{IPCE_2024_CVPR,
    author    = {Peng, Fei and Fu, Huiyuan and Ming, Anlong and Wang, Chuanming and Ma, Huadong and He, Shuai and Dou, Zifei and Chen, Shu},
    title     = {AIGC Image Quality Assessment via Image-Prompt Correspondence},
    booktitle = {Proceedings of the IEEE/CVF Conference on Computer Vision and Pattern Recognition (CVPR) Workshops},
    month     = {June},
    year      = {2024},
    pages     = {6432-6441}
}

@inproceedings{MA-AGIQA,
  title={Large Multi-modality Model Assisted AI-Generated Image Quality Assessment},
  author={Wang, Puyi and Sun, Wei and Zhang, Zicheng and Jia, Jun and Jiang, Yanwei and Zhang, Zhichao and Min, Xiongkuo and Zhai, Guangtao},
  booktitle={Proceedings of the 32nd ACM International Conference on Multimedia},
  pages={7803--7812},
  year={2024}
}

@inproceedings{AIGIQA20K,
  title={AIGIQA-20K: A Large Database for AI-Generated Image Quality Assessment},
  author={Chunyi Li and Tengchuan Kou and Yixuan Gao and Yu Shan Cao and Wei Sun and Zicheng Zhang and Yingjie Zhou and Zhichao Zhang and Weixia Zhang and Haoning Wu and Xiaohong Liu and Xiongkuo Min and Guangtao Zhai},
  year={2024},
  url={https://api.semanticscholar.org/CorpusID:268889245}
}

@article{AGIQA1K,
  title={A Perceptual Quality Assessment Exploration for AIGC Images},
  author={Zhang, Zicheng and Li, Chunyi and Sun, Wei and Liu, Xiaohong and Min, Xiongkuo and Zhai, Guangtao},
  journal={arXiv preprint arXiv:2303.12618},
  year={2023}
}

@ARTICLE{AGIQA-3K,
  author={Li, Chunyi and Zhang, Zicheng and Wu, Haoning and Sun, Wei and Min, Xiongkuo and Liu, Xiaohong and Zhai, Guangtao and Lin, Weisi},
  journal={IEEE Transactions on Circuits and Systems for Video Technology}, 
  title={AGIQA-3K: An Open Database for AI-Generated Image Quality Assessment}, 
  year={2023},
  pages={1-1},
  doi={10.1109/TCSVT.2023.3319020}}

@article{aigciqa2023,
  title={Aigciqa2023: A large-scale image quality assessment database for ai generated images: from the perspectives of quality, authenticity and correspondence},
  author={Wang, Jiarui and Duan, Huiyu and Liu, Jing and Chen, Shi and Min, Xiongkuo and Zhai, Guangtao},
  journal={arXiv preprint arXiv:2307.00211},
  year={2023}
}

@article{PKUAIGIQA4K,
  title={PKU-AIGIQA-4K: A Perceptual Quality Assessment Database for Both Text-to-Image and Image-to-Image AI-Generated Images},
  author={Jiquan Yuan and Fanyi Yang and Jihe Li and Xinyan Cao and Jinming Che and Jinlong Lin and Xixin Cao},
  journal={ArXiv},
  year={2024},
  volume={abs/2404.18409},
  url={https://api.semanticscholar.org/CorpusID:269449873}
}

@inproceedings{CLIPIQA,
  title={Exploring clip for assessing the look and feel of images},
  author={Wang, Jianyi and Chan, Kelvin CK and Loy, Chen Change},
  booktitle={Proceedings of the AAAI Conference on Artificial Intelligence},
  volume={37},
  number={2},
  pages={2555--2563},
  year={2023}
}

@article{kingma2014adam,
  title={Adam: A method for stochastic optimization},
  author={Kingma, Diederik P and Ba, Jimmy},
  journal={arXiv preprint arXiv:1412.6980},
  year={2014}
}

@misc{IP-IQA,
      title={Bringing Textual Prompt to AI-Generated Image Quality Assessment}, 
      author={Bowen Qu and Haohui Li and Wei Gao},
      year={2024},
      eprint={2403.18714},
      archivePrefix={arXiv},
      primaryClass={cs.CV},
      url={https://arxiv.org/abs/2403.18714}, 
}

@inproceedings{tang2025clip,
  title={CLIP-AGIQA: Boosting the Performance of AI-Generated Image Quality Assessment with CLIP},
  author={Tang, Zhenchen and Wang, Zichuan and Peng, Bo and Dong, Jing},
  booktitle={International Conference on Pattern Recognition},
  pages={48--61},
  year={2025},
  organization={Springer}
}

\end{document}